\documentclass[11pt]{article}
\usepackage{coling2020}
\usepackage{times}
\usepackage{url}
\usepackage{latexsym}
\usepackage{amsthm}
\usepackage{amsmath}
\usepackage{amssymb}
\usepackage{array,multirow,graphicx}
 
\usepackage{bm}
 
\usepackage{booktabs}
 
\usepackage{dsfont}

\usepackage{multirow}
\usepackage{tabularx}
\usepackage[para,online,flushleft]{threeparttable}

\usepackage{hyperref}

\usepackage{wrapfig}

\newcolumntype{H}{>{\hsize=.5\hsize}X}
\newcolumntype{K}{>{\hsize=2\hsize}X}

\newcolumntype{S}[1]{>{\raggedright\let\newline\\\arraybackslash\hspace{0pt}}H{#1}}
\newcolumntype{B}[1]{>{\raggedright\let\newline\\\arraybackslash\hspace{0pt}}X{#1}}
\newcolumntype{Z}[1]{>{\raggedright\let\newline\\\arraybackslash\hspace{0pt}}K{#1}}

\usepackage{rotating}

\newcommand*{\affaddr}[1]{#1} 
\newcommand*{\affmark}[1][*]{\textsuperscript{#1}}
\newcommand*{\email}[1]{\texttt{#1}}

\usepackage{subfig}
\usepackage{graphicx}

\graphicspath{{pictures/}}

\colingfinalcopy 

\title{Static Fuzzy Bag-of-Words: a lightweight sentence embedding algorithm}

\author{%
Matteo Muffo\affmark[$\dag$ *]{\normalfont ,} Roberto Tedesco\affmark[$\dag$]{\normalfont ,} Licia Sbattella\affmark[$\dag$] \and Vincenzo Scotti\affmark[$\dag$]\\
\affaddr{\affmark[$\dag$]DEIB, Politecnico di Milano \\ Via Golgi 42, 20133, Milano (MI), Italy}\\
\affaddr{\affmark[*]Indigo.ai \\ Via Torino 61, 20123, Milano (MI), Italy}\\
\email{matteo.muffo@mail.polimi.it} \qquad \email{roberto.tedesco@polimi.it} \\
\email{licia.sbattella@polimi.it} \qquad \email{vincenzo.scotti@polimi.it} \\
}

\date{}

\begin{document}
\maketitle
\begin{abstract}
The introduction of embedding techniques has pushed forward significantly the Natural Language Processing field.
Many of the proposed solutions have been presented for word-level encoding; anyhow, in the last years, new mechanism to treat information at an higher level of aggregation, like at sentence- and document-level, have emerged.
With this work we address specifically the sentence embeddings problem, presenting the \emph{Static Fuzzy Bag-of-Word} model. Our model is a refinement of the Fuzzy Bag-of-Words approach, providing sentence embeddings with a predefined dimension. SFBoW provides competitive performances in Semantic Textual Similarity benchmarks, while requiring low computational resources.
\end{abstract}

\section{Introduction}
Natural Language Processing (NLP) has gained a lot of traction in the last years, especially thanks to the introduction of learnt semantic representations.
Those representations (usually) called embeddings are, in the textual context, real-valued vectors able to represent the semantic meaning of words, sentences or even documents, in a Euclidean space.
These vectors are features generated by models trained using a \emph{self-supervised} approach on a huge corpus of unlabelled text.
According to \newcite{6472238}, leveraging features obtained through a self-supervised approach instead of ``hand-selected'' features it's of crucial importance for NLP.

Textual learnt representations immediately turned out to be significative in many NLP tasks, from more simple ones, like Part-Of-Speech (POS) tagging, Named Entity Recognition (NER) and language modelling \cite{collobert2007fast,collobert2008unified,collobert2011nlpscratch}, to more complex problems such as Machine Translation \cite{sutskever2014sequence}, and even Conversational Systems \cite{sordoni-etal-2015-neural}.
In fact, this representation significantly moved forward the state of the art.

Results in the aforementioned tasks have been boosted mostly thanks to learnt word-level semantic vectors, i.e. \emph{word embeddings}; however, as pointed out by \newcite{he2014deep}, for many problems (like web search, question answering and image captioning) having access to higher level representations is crucial: this is where \emph{sentence embeddings} find their usefulness.

Recent outcomes show that \emph{contextual} representations, learnt through Transformer Networks \cite{vaswani2017attention}, provide better performances in all those tasks and are slowly substituting ``static'' embeddings.
Even though the results of these models are remarkable, their usability is strongly restricted because of their high demand in terms of computational resources, so we decided to focus on more lightweight solutions.

With this work, we introduce the Static Fuzzy Bag-of-Words (SFBoW) model, an improvement of the work from \newcite{zhelezniak2019dont}, for sentence embedding.
SFBoW is characterised by interesting results in Semantic Textual Similarity (STS), without the necessity of high computational power.

The rest of this document is organised as follows: Section~\ref{sec:relatedworks} presents the related works in the field of learnt semantic representations, Section~\ref{sec:sfbowmodel} introduces our SFBoW model, Section~\ref{sec:experiments} describes the planned experiments to assess the quality of our model, Section~\ref{sec:results} presents the results of experiments and, finally, Section~\ref{sec:conclusionsandfutureworks} sums up the entire work. 

\section{Related works}
\label{sec:relatedworks}
Our work revolves around the concept of \emph{vector semantics}: the idea that the meaning of a word or a sentence can be modelled as a vector \cite{osgood1957measurement}.

First steps on this subject were made in Information Retrieval (IR) context with the vector space model \cite{salton1971smart,10.5555/22908}, where documents and queries were represented as high dimensional (vocabulary size)  sparse embedding vectors.
In this model, each dimension is used to represent a word, so that given a vocabulary $\mathcal{V}$:
\begin{itemize}
\setlength\itemsep{0em}
\item A word $w_i \in \mathcal{V}$, with $i \in \left[1, | \mathcal{V} |\right] \subseteq \mathbb{N}$, is expressed as a so called ``one hot'' binary vector $\mathbf{v}_{w_i} \in \mathds{1}^{|\mathcal{V}|}$, where, calling $v_{w_i,j}$ the $j$-th element of the word vector, it holds that $v_{w_i, j} = 1 \Longleftrightarrow j = i$;
\item A sentence $S$ is expressed as vector $\bm{\mu}_S \in \mathbb{N}^{|\mathcal{V}|}$, where $\mu_{S,i}$, the $i$-th element of vector $\bm{\mu}_S$, represents the number of times, $c_{S,i}$, word $w_i$ appears in sentence $S$.
\end{itemize}
The sentence representation, used also for text documents, is called Bag-of-Words (BoW), and can be summarised in the following equation:
\begin{equation}
\bm{\mu}_S = \sum_{i = 1}^{ | \mathcal{V} |} c_{S,i} \cdot \mathbf{v}_{w_i}
\end{equation}

As pointed out by \newcite{bengio2003neural}, these embedding models needed to be replaced because of their sparsity, which made them time/memory consuming, and the induced orthogonality among vectors with similar meanings.

\subsection{Word and sentence embeddings: a brief summary}
\label{subsec:wasebriefsumm}
Word embeddings refer to the dense semantic vector representation of words.
Approaches can be divided into: \emph{prediction-based} and \emph{count-based} \cite{baroni-etal-2014-dont}.

The former group identifies the embeddings obtained through the training of models for next/missing word prediction given a context,  it was started by \newcite{bengio2003neural} and encompasses models like Word2Vec \cite{mikolov2013efficient,mikolov2013distributed} and FastText \newcite{fasttext}.
The latter group refers to the embeddings obtained leveraging words co-occurrence counts in a corpus, it has a longer history, starting from Latent Semantic Indexing (LSI) over term-document matrices \cite{deerwester1989computer} up to more recent solutions like GloVe from \newcite{pennington2014glove}.

Among the aforementioned models, there are some (like Word2Vec, GloVe or FastText) belonging to a class called \emph{shallow} models, where the embedding of a word $w_i$ can be simply extracted through lookup over the rows of the embedding matrix $\mathbf{W} \in \mathbb{R}^{|\mathcal{V}| \times d}$, with $d$ being the desired dimensionality of the embedding space.
Given the word (column) vector $\mathbf{v}_{w_i}$, the corresponding word embedding $\mathbf{u}_{w_i} \in \mathbb{R}^d$ can be computed as (see Section~\ref{subsec:fbow}):
\begin{equation}
\label{eq:wedef}
\mathbf{u}_{w_i} = \mathbf{W}^\top \cdot \mathbf{v}_{w_i}
\end{equation}

More recently, the introduction of transformer-based Language Models (LMs), like BERT \cite{devlin2018bert} (and all its variants) or the two versions of GPT \cite{radford2018improving,radford2019language}, has spread the concept of \emph{contextual embeddings}; such embeddings proved to be particularly helpful for a wide variety of NLP problems, as shown by the leader-boards of many NLP benchmarks like SuperGLUE \cite{wang2019superglue}.

The inherent hierarchical structure of the human language makes it hard to understand a text from single words; thus, the birth of higher level semantic representations for sentences, which are the sentence embeddings, was just a natural consequence.
As for the Word embeddings, also sentence embeddings are organised into two groups: \emph{parametrised} and \emph{non-parametrised}, depending on whether the model requires parameter training or not.

First examples of parametric models come from \newcite{kiros2015skip}, with the Skip-Thoughts vectors, later followed by Sent2Vec \cite{Pagliardini_2018}, which generalised Word2Vec.
Transformer LMs influenced the state of the art also at this level, a clear example is given by Sentence-BERT \cite{reimers2019sentencebert}, obtained by fine-tuning on two Natural Language Inference (NLI) corpora: the Stanford NLI \cite{snli} and the Multi-Genre NLI \cite{snli2}.

Non-parametric models, instead, show that simply aggregating the information from pre-trained word embeddings, for example through averaging \cite{arora2019simple,yang2019parameter}, is sufficient to represent represent higher-level entities like sentences and paragraphs.

All these models rely on the assumption that cosine similarity is the correct metric to compute ``meaning distance'' between sentences, and this is why parametric models are trained specifically to minimise this measure for similar sentences and maximise it for dissimilar sentences.
However this may not be the only and best measure. 
In fact \newcite{zhelezniak2019dont} proposed to follow a fuzzy set representation of sentences and to rely on fuzzy Jaccard similarity, instead of the cosine one.
As a result, their DynaMax model outperformed many non-parametric models and performed comparably to parametric ones under cosine similarity measurements, even if competitors were trained directly to optimise that metric, while DynaMax approach was completely unrelated to that objective.

The use of fuzzy sets to represent documents is not new, in fact it was already proposed by \newcite{zhao_e_mao}.
However, they relied on a different approach to compute fuzzy membership, with respect to \newcite{zhelezniak2019dont}, which lead to inferior results.

\subsection{Sentence embeddings: Fuzzy Bag-of-Words and DynaMax model}
\label{subsec:fbow}
The Fuzzy Bag-of-Words (FBoW) model for text representation, proposed by \newcite{zhao_e_mao}, then generalised and improved by \newcite{zhelezniak2019dont} with their DynaMax through the introduction of a better similarity metric, represent the starting point of our work, which will be described in Section~\ref{sec:sfbowmodel}.

The BoW approach, described at the beginning of Section~\ref{sec:relatedworks}, can be seen as a multi-set representation of text.
As such it enables to measure similarity between two sentences with set similarity measures, like Jaccard, Otsuka and Dice indexes.
These indexes, as \newcite{zhelezniak2019dont} pointed out, share all a common pattern to measure the similarity $\sigma$ between two sets $A$ and $B$:
\begin{equation}
\sigma\left(A,B\right) = n_{\textit{shared}}\left(A,B\right) / n_{\textit{total}}\left(A,B\right)
\end{equation}
Where $n_{\textit{shared}}\left(A,B\right)$ denotes the count of shared elements and $n_{\textit{total}}\left(A,B\right)$ denotes the count of total elements.
In particular, the Jaccard index is defined as:
\begin{equation}
\label{eq:jaccardindex}
\sigma_{\textit{Jaccard}}\left(A, B\right) = \left| A \cap B \right| / \left| A \cup B \right|
\end{equation}

Simple set similarity is however a rigid approach as it allows for some degree of similarity when the very same words appear in both sentences, but fails in presence of synonyms.
This is where \emph{Fuzzy Sets theory} \cite{zadeh} comes handy: in fact, fuzzy sets enable to interpret each word in $\mathcal{V}$ as a singleton set and measure the degree of membership of any word to this one-element set as the similarity between the two considered words \cite{zhao_e_mao}.

In their FBoW model, \newcite{zhao_e_mao} proposed to work exactly in this way:
\begin{itemize}
\setlength\itemsep{0em}
\item each single word $w_i$ is interpreted as a singleton $\left\{w_i\right\}$; thus, the membership degree of any word $w_j$ in the vocabulary (with $j \in \left[1, | \mathcal{V} |\right] \subseteq \mathbb{N}$) with respect to this set is computed as the similarity $\sigma$ between $w_i$ and $w_j$. These similarities can be used to fill a $|\mathcal{V}|$-sized vector $\hat{\textbf{v}}_{w_i}$ used to provide the fuzzy representation of $w_i$ (the $j$-th element $\hat{\textbf{v}}_{w_i,j}$ being $\sigma\left(w_i, w_j\right)$);
\item a sentence $S$ is simply defined through the fuzzy union operator, which is determined by the $\max$ operator over the membership degrees. In this case the $S$ is represented by a vector of $| \mathcal{V} |$ elements.
\end{itemize}

Then, to reduce the dimension of the vector for $S$, the \newcite{zhelezniak2019dont} generalised the FBoW approach, computing the \emph{fuzzy embedding} of a word singleton as:
\begin{equation}
\hat{\textbf{v}}_{w_i} = \mathbf{U} \cdot \mathbf{u}_{w_i} = \mathbf{U} \cdot \mathbf{W}^\top \cdot \mathbf{v}_{w_i}
\end{equation}
where, $\mathbf{W} \in \mathbb{R}^{|\mathcal{V}| \times d}$ is a word embedding matrix (defined as in Section~\ref{subsec:wasebriefsumm}), $\mathbf{u}_{w_i}$ is defined in Equation~\ref{eq:wedef} and $\mathbf{U} \in \mathbb{R}^{u \times d}$ (with $u$ being the desired dimension of the fuzzy embeddings) is the \emph{universe matrix}, derived from the \emph{universe set} $U$, which is defined as ``the set of all possible terms that occur in a certain domain''. The generalised FBoW produces vectors of $u$ elements, where $u=|U|$.


Finally, given the fuzzy embeddings of the words in a sentence $S$, the generalised FBoW representation of $S$ is a vector $\hat{\bm{\mu}}_S$ whose $j$-th element $\hat{\mu}_{S,j}$ ($j \in \left[1,u\right] \subseteq \mathbb{N}$) can be computed as:
\begin{equation}
\hat{\mu}_{S,j} = \max_{w_i \in S} c_{S,i} \cdot \hat{v}_{w_i, j}
\end{equation}
where $c_{S,i}$ and $ \hat{v}_{w_i, j}$ are, respectively, the number of occurrences of word $w_i$ in sentence $S$ and the $j$-th element of the $\hat{\textbf{v}}_{w_i}$ vector. 

The universe set can be defined in different ways; in fact, \newcite{zhelezniak2019dont} suggested some alternatives for the choice of the universe set, and hence the universe matrix.
Among the proposed solutions, in their best performing algorithm for fuzzy sentence embeddings, the DynaMax, they choose to build the universe matrix from the word embedding matrix, stacking solely the embedding vectors of the words appearing in the sentences to be compared.

Notice that in this way the resulting universe matrix is not unique, as a consequence neither are the embeddings.
This condition can be noticed from the description of the algorithm and from the definition of the universe matrix: when comparing two sentences $S_a$ and $S_b$, the universe set $U$ used in their comparison is $U \equiv S_a \cup S_b$, so the resulting sentence embeddings have size  $u = \left|U\right| = \left|S_a \cup S_b\right|$. In fact, the universe matrix is given by
\begin{equation}
\mathbf{U} = \begin{bmatrix}\mathbf{u}_{w_i} \forall w_i \in U \end{bmatrix}^\top
\end{equation}

This characteristic is unfortunate as, for example in the field of IR, it requires a complete re-encoding of the entire document achieve for each query.

The real improvement of DynaMax is in the introduction of the fuzzy Jaccard index to compute the semantic similarity between two sentences $S_a$ and $S_b$ (see Equation~\ref{eq:fuzzyjaccardindex}), rather than the generalisation of the FBoW, which replaced the cosine similarity employed by \newcite{zhao_e_mao}. 
\begin{equation}
\label{eq:fuzzyjaccardindex}
\hat{\sigma}_{\textit{Jaccard}}\left(\hat{\bm{\mu}}_{S_a}, \hat{\bm{\mu}}_{S_b}\right) = \frac{\sum_{i=1}^u \min \left(\hat{\mu}_{S_a,i}, \hat{\mu}_{S_b,i}\right)}{\sum_{i=1}^u \max \left(\hat{\mu}_{S_a,i}, \hat{\mu}_{S_b,i}\right)}
\end{equation}


\section{Static Fuzzy Bag-of-Words model}
\label{sec:sfbowmodel}

\begin{figure}[b]
\begin{center}
\includegraphics[width=.8\textwidth]{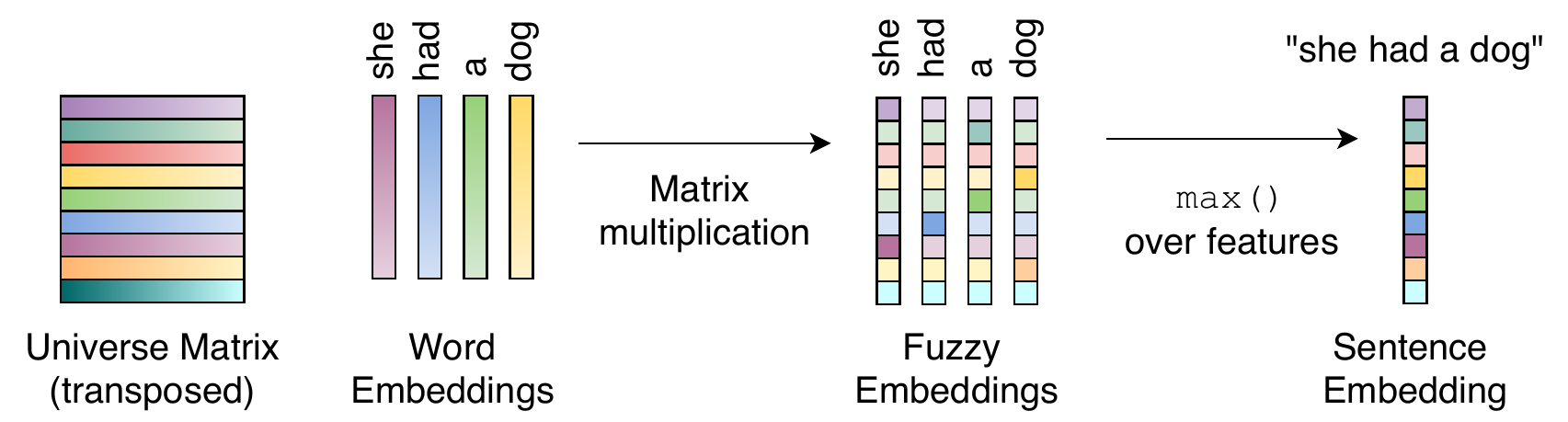}
\caption{Visualization of the Sentence Embedding computation process using SFBoW.}
\label{fig:sevisual}
\end{center}
\vspace{-5mm}
\end{figure}

Starting from the the DynaMax, evolved from the FBoW model, we developed our follow up aimed at providing a unique matrix $\mathbf{U}$ and thus embeddings with a predefined dimension.
In Figure~\ref{fig:sevisual} is represented the visualization of our approach.

\subsection{Concerning the word embeddings}
\label{sec:subsets}
Word embeddings play a central role in our algorithm as they also provide the start point of the construction of the universe matrix.
For the scope of this work, we leverage pre-trained shallow models (more details in Section~\ref{sec:exploredpossibilities}) for two main reasons:
\begin{itemize}
\setlength\itemsep{0em}
\item the model is encoded in a matrix where each row corresponds to a word;
\item we want to provide a sentence embedding that doesn't require training, easing its accessibility.
\end{itemize}

The vocabulary of these models, composed starting from all the tokens in the training corpora, is usually bigger than the English vocabulary, as it contains named entities, incorrectly spelled words, non-existing words, URLs, email addresses, etc.
To reduce the computational effort needed for construction and usage of the universe matrix, we have considered some subsets of the employed word embedding model's vocabulary.
Depending on the experiment, we work with either the 100000 most frequently used terms, the 50000 most frequently used terms (terms frequencies are given by the corpora used to train the word embedding model) or the subset composed of all the spell-checked terms present in a reference English dictionary (obtained through the Aspell English spell-checker\footnote{\url{http://aspell.net}}).

In the following sections, the $\check{\mathbf{W}}$ symbol refers to these as \emph{reduced} word embedding matrices/models.

\subsection{Universe matrix}
\label{sec:univmat}
During the experiments, we try three main approaches to build the universe matrix $\mathbf{U}$: the first two -- proposed, but not explored, by the original authors of DynaMax -- consist, respectively, in the usage of clustered embedding matrix and an identity matrix with the rank equal to the size of the word embeddings.
The last approach, instead, consists in the application of a multivariate analysis techniques to the word embedding matrix to build the universe one.
In the following formulae we refer to $d$ as the dimensionality of the word embedding vectors, while the SFBoW embedding of the singleton of word $w_i$ is represented as $\check{\textbf{v}}_{w_i}$.

\paragraph{Clustering} The idea is to group the embedding vectors into clusters, and use their centroids; in this way the fuzzy membership will be computed over the clusters -- which are expected to host semantically similar words -- instead of all the word singletons.
The universe set is thus built out of abstract entities only, which are the centroids.
Considering $k$ cluster centroids, the universe matrix $\mathbf{U} = \mathbf{K}^\top \in \mathbb{R}^{k \times d}$, and thus SFBoW $k$-dimensional embedding $\check{\textbf{v}}_{w_i}$ of the singleton of word $w_i$ is
\begin{equation}
\check{\textbf{v}}_{w_i} = \mathbf{K}^\top \cdot \mathbf{u}_{w_i} = \begin{bmatrix}\mathbf{k}_1, \ldots, \mathbf{k}_k\end{bmatrix}^\top \cdot \mathbf{u}_{w_i} = \mathbf{K}^\top \cdot \mathbf{W}^\top \cdot \mathbf{v}_{w_i}
\end{equation}
where $\mathbf{k}_j$, the $j$-th (with $j \in \left[1,k\right] \subseteq \mathbb{N}$) column of $\mathbf{K}$, corresponds to the centroid of the $j$-th cluster. This approach generates $k$-dimensional word embeddings and sentence embeddings.

\paragraph{Identity} Alternatively, instead of looking for a group of semantically similar words that may possibly form a significant group, useful for semantic similarity, we consider the possibility of re-using the word embedding dimensions (features) to represent the semantic content of a sentence.
So, we just use the identity matrix as the universe: $\mathbf{U} = \mathbf{I}$, with $\left|\mathbf{I}\right| = d \times d$, so that $\check{\textbf{v}}_{w_i} \in \mathbb{R}^d$ is 
\begin{equation}
\check{\textbf{v}}_{w_i} = \mathbf{I} \cdot \mathbf{u}_{w_i} = \mathbf{I} \cdot \mathbf{W}^\top \cdot \mathbf{v}_{w_i}
\end{equation}
This approach generates $d$-dimensional word embeddings and sentence embeddings.

\paragraph{Multivariate analysis} The same idea moves our multivariate analysis proposal. In fact, judging by previous results, word embeddings aggregated in the correct way might be sufficient to provide a semantically valid representation of the sentence they compose.
What can bring better results, might be as simple as roto-translate the reference system of the embedding representation.
In this sense, we propose to use to compute the fuzzy membership, and hence the fuzzy Jaccard similarity index, over these dimensions resulting from roto-translation, expecting that this ``new perspective'' will expose better the semantic content.
So, defining $ \mathbf{U} =  \mathbf{M}$, where $\mathbf{M}$ is the transformation matrix, with $\left|\mathbf{M}\right| = d \times d$, we have that $\check{\textbf{v}}_{w_i} \in \mathbb{R}^d$ is 
\begin{equation}
\check{\textbf{v}}_{w_i} = \mathbf{M} \cdot \mathbf{u}_{w_i} = \mathbf{M} \cdot \mathbf{W}^\top \cdot \mathbf{v}_{w_i}
\end{equation}
This approach generates $d$-dimensional word embeddings and sentence embeddings.
\\

Clustering and multivariate analysis can be applied to the whole embedding vocabulary or the subsets of the vocabulary introduced in Section~\ref{sec:subsets}.
A part from reducing the computational time, we do so also to see if these subsets are sufficient to provide a useful representation.

\section{Experiments with the Static Fuzzy Bag-of-Words}
\label{sec:experiments}
In order to find the best solution in terms of word embedding matrix and universe matrix, we explored various possibilities.
Then, to measure the goodness of our sentence embeddings, we leveraged a series of STS tasks and compared the results with preceding models.

\subsection{Word embeddings}
\label{sec:exploredpossibilities}
For what concerns the word embeddings, we have decided to work with a selection of four models:
\begin{itemize}
\setlength\itemsep{0em}
\item Word2Vec, with $300$-dimensional embeddings \cite{mikolov2013distributed};
\item GloVe, with $300$-dimensional embeddings \cite{pennington2014glove};
\item FastText, with $300$-dimensional embeddings \cite{fasttext};
\item Sent2Vec, with $700$-dimensional embeddings \cite{Pagliardini_2018}.
\end{itemize}
As shown by the word embedding models list, we are employing also a Sent2Vec sentence embedding model. 
In fact the embedding matrix of this model can be used for word embeddings too \cite{sent2vec_word}.
During the experiments we focused on the universe matrix construction, for this reason we relied on pre-trained models for word embeddings, available on the web.

\subsection{Universe matrices}
The universe matrices we considered, are divided in three buckets, as described in Section~\ref{sec:univmat}.

\paragraph{Clustering}
Universe matrices built using clustering leverage three different algorithms: k-Means \cite{macqueen1967}, Spherical k-Means \cite{hornik2012spherical} and DBSCAN \cite{Ester96adensity-based}.

We selected k-Means and Spherical k-Means because they usually lead to good clustering results, the latter was specifically designed for textual purposes, with a low demand in terms of time and computation resources.
For all algorithms we considered the same values for $k$ (the number of centroids), which were 100, 1000, 10 000 and 25 000.
For all the values of $k$, we performed clustering on different subsets of the vocabulary: k-Means was applied on the whole English vocabulary as well as to the top 100 000 frequently used words subset, while Spherical k-Means was applied to the subset of the first 50 000 frequently used words (in order to reduce computational time).

We explored also a density based algorithm (DBSCAN), which does not require to define in advance the number of clusters, using euclidean and cosine distance between the word embedding.
For the former case we varied the radius of the neighbourhood $\varepsilon$ between $3$ and $8$ and worked over the same two subsets considered for k-Means, for the latter $\varepsilon$ was between $0.1$ and $0.55$ and it was applied over the subset of the first 50 000 frequently used words (for computational reasons, as we did for Spherical k-Means).

\paragraph{Identity}
This approach consists in using the identity matrix as universe, in this way the singletons we use to compute the fuzzy membership are the dimensions of the word embeddings, which corresponds to the learnt features.
This is the most lightweight method as is just requires to compute the word embeddings of a sentence and then the fuzzy membership over the same $d$ dimensions.

\paragraph{Multivariate analysis}
We adopted the Principal Component Analysis (PCA) to get a rotation matrix which will serve as universe matrix to the SFBoW.
In fact through PCA the $d$-dimensional word embedding vectors are decomposed along the $d$ orthogonal directions of their variance.
These components are then reordered in decreasing order of explained variance and used to represent our semantic fuzzy sets.

The principal component of the reduced word embedding matrix $\check{\mathbf{W}}$ are described by the matrix $\mathbf{T} = \mathbf{P}^\top \cdot \check{\mathbf{W}}$, where $\mathbf{P}$ is a $d \times d$ matrix whose columns are the eigenvectors of the matrix $\check{\mathbf{W}}^\top \cdot \check{\mathbf{W}}$.
With our approach, the matrix $\mathbf{P}^\top$, sometimes called the \emph{whitening} or \emph{sphering transformation matrix}, serves as universe matrix $\mathbf{U}$.
In this way, the SFBoW embedding of a word singleton becomes
\begin{equation}
\check{\textbf{v}}_{w_i} = \mathbf{P}^\top \cdot \mathbf{u}_{w_i} = \mathbf{P}^\top \cdot \check{\mathbf{W}}^\top \cdot \mathbf{v}_{w_i}
\end{equation}

As for the clustering approach, we experimented with both the whole vocabulary and the top 100 000 frequently used words.

\subsection{Data}
The evaluation of our SFBoW is done through a series of reference benchmarks; we selected the STS benchmark series, one of the tasks of the International Workshop on Semantic Evaluation (SemEval)\footnote{\url{https://aclweb.org/aclwiki/SemEval_Portal}}.


SemEval is an ongoing series of evaluations on computational semantics; among these evaluations, the STS benchmark has become a reference for scoring of sentence embedding algorithms.
In fact, all the previous models we are considering for a comparison have been benched against STS; this is because the benchmark highlights a model capability to provide a meaningful semantic representation by scoring the correlation between model's and human's judgements.
For this reason, and also to make comparisons possible, we decided to evaluate SFBoW on STS.

We worked only on the English language, using the editions of STS from 2012 to 2016 \cite{sts12,sts13,sts14,sts15,sts16}.
Each year, a collection of corpora coming from different sources has been created and manually labelled, in Table~\ref{tab:supportcorpora} is possible to have a reference in terms of support for each edition.
Thanks to the high number of samples, we are confident about the robustness of our results.

\begin{table}[t]
\caption{Support of the corpora of the STS benchmark series.}
\begin{center}
\begin{tabular}{l | ccccc | r}
\toprule
Edition & STS '12 & STS '13 & STS '14 & STS '15 & STS '16 & Total \\ 
\midrule
Number of sentence pairs & 5250 & 2250 & 3750 & 3000 & 1186 & 15 436 \\
\bottomrule
\end{tabular}
\end{center}
\label{tab:supportcorpora}
\vspace{-5mm}
\end{table}

The samples constituting the corpora are pair of sentences with a human-given similarity score (the \emph{gold labels}).
The provided score is a real valued index obtained averaging those of multiple crowd-sourced workers, and is scaled in a $\left[0,1\right] \in \mathbb{R}$ interval.
The final goal of our work is to provide a model able to provide a score as close as possible to that of the humans.

\subsection{Evaluation}
To assess the quality of our model, we used it to compute the similarity score between the sentence pairs provided by the five tasks, and we compared the output with the target labels.
The results are computed as the correlation between the similarity score produced by SFBoW and the human one, using the Spearman's $\rho$ measure, as suggested by \newcite{reimers-etal-2016-task}.
The index SFBoW employs to compute word similarity is the fuzzy Jaccard similarity index, as suggested by \newcite{zhelezniak2019dont}.

To have terms of comparison, we establish a baseline through the simplest models possible, the average of the word embedding in a sentence, leveraging three different word embedding models: Word2Vec, GloVe and FastText.  
We also provide results from more complex models: SIF weighting (applied to GloVe), Sent2Vec, DynaMax (built using Word2Vec, GloVe and FastText) and Sentence-BERT.

All the sentence embedding models, except DynaMax, and the baselines are scored using the results of cosine similarity; DynaMax scores are instead obtained using the fuzzy Jaccard similarity index.

\section{Results}
\label{sec:results}
The results of the Spearman's $\rho$ correlation in the STS benchmark of our SFBoW are reported in the last three rows of Table~\ref{tab:resultscomparison}.
The reported values belong to the FSBoW configurations that achieved a best score, among the variants we considered for the experiments, in at least one task; if interested in the complete experimental results reporting all the SFBoW configurations, please refer to Appendix~\hyperref[sec:extendedresults]{A}.

\begin{table}[ht]
\caption{Comparison in the results over the STS benchmark. The last block shows the best performing SFBoW models we evaluated. All averages are expressed as: $\textit{average score} \pm \textit{standard deviation}$. Bold values represent the best score in each column, underlined values are the second best score in the column.}
\begin{center}
\begin{threeparttable}
\begin{tabular}{l | ccccc | c | c}
\toprule
\multirow{3}{*}{} & \multicolumn{7}{c}{Results (Spearman's $\rho$)} \\ \cline{2-8}
& \multicolumn{5}{c |}{STS} & \multirow{2}{*}{Average} & \multirow{2}{*}{Weighted average} \\ \cline{2-6}
& '12 & '13 & '14 & '15 & '16 & & \\
\midrule
Word2Vec~\tnote{a} & $55.46$ & $58.23$ & $64.05$ & $67.97$ & $66.28$ & $62.4\pm4.78$ & $61.21\pm5.04$ \\
GloVe~\tnote{a} & $53.28$ & $50.76$ & $55.63$ & $59.22$ & $57.88$ & $55.35\pm3.06$ & $54.99\pm2.80$ \\
FastText~\tnote{a} & $58.82$ & $58.83$ & $63.42$ & $69.05$ & $68.24$ & $63.67\pm4.40$ & $62.65\pm4.20$ \\
\midrule
SIF weighting~\tnote{b} & $56.04$ & $\underline{62.74}$ & $64.29$ & $69.89$ & $70.71$ & $64.73\pm5.33$ & $62.84\pm5.54$ \\
Sent2Vec & $56.26$ & $57.02$ & $65.82$ & $74.46$ & $69.01$ & $64.51\pm7.00$ & $63.21\pm7.13$ \\
DynaMax~\tnote{c} & $55.95$ & $60.17$ & $65.32$ & $73.93$ & $71.46$ & $65.37\pm6.73$ & $63.53\pm6.92$ \\
DynaMax~\tnote{b} & $57.62$ & $55.18$ & $63.56$ & $70.40$ & $71.36$ & $63.62\pm6.53$ & $62.25\pm5.85$ \\
DynaMax~\tnote{d} & $61.32$ & $61.71$ & $66.87$ & $\underline{76.51}$ & $\underline{74.71}$ & $\underline{68.22}\pm6.37$ & $\underline{66.71}\pm6.1$\\
Sentence-BERT & $\mathbf{72.27}$ & $\mathbf{78.46}$ & $\mathbf{74.90}$ & $\mathbf{80.99}$ & $\mathbf{76.25}$ & $\mathbf{76.57}\pm2.98$ & $\mathbf{75.81}\pm3.27$ \\
\midrule
SFBoW~\tnote{d, e, f} & $61.31$ & $51.21$ & $\underline{67.47}$ & $72.90$ & $73.88$ & $65.35\pm8.37$ & $64.55\pm7.20$ \\
SFBoW~\tnote{d, g, h} & $\underline{61.42}$ & $51.36$ & $66.44$ & $72.74$ & $73.72$ & $65.14\pm8.21$ & $64.32\pm7.00$ \\
SFBoW~\tnote{d, g, i} & $60.03$ & $51.96$ & $66.36$ & $72.39$ & $73.25$ & $64.80\pm7.99$ & $63.81\pm6.93$ \\
\bottomrule
\end{tabular}
\begin{tablenotes}
\item[a] Used as baseline.
\item[b] Built upon a GloVe model for word embeddings.
\item[c] Built upon a Word2Vec model for word embeddings.
\item[d] Built upon a FastText model for word embeddings.
\item[e] Best average score.
\item[f] Universe matrix is the identity matrix.
\item[g] Universe matrix is the PCA projection matrix.
\item[h] Universe matrix is built from the English vocabulary.
\item[i] Universe matrix is built from the top 100~000 most frequent words.
\end{tablenotes}
\end{threeparttable}
\end{center}
\label{tab:resultscomparison}
\vspace{-5mm}
\end{table}%

Among the four word embeddings models we tried, the best performing is FastText, confirming the results of  \newcite{zhelezniak2019dont}.
Concerning the choice of the universe matrix, instead, the best scores are achieved either with Identity matrix or with PCA rotation matrix, highlighting how the features described by the word embeddings provide a better description of the semantic content of sentences.

About the choice of the universe matrix, clustering provided poor results, so Table~\ref{tab:resultscomparison} reports only the scores from Identity matrix and PCA.
Density based clustering turned out to give meaningless results, for this reason its analysis is omitted.
k-Means clustering, instead, gave more promising results, but still inadequate if compared to Identity matrix and PCA.
The analysis of Within Clusters Sum-of-Squares (WCSoS) and Between Clusters Sum-of-Squares (BCSoS) showed that there's still room for improvement so, as a future work, we plan to further investigate this topic.

\begin{figure}[t]
\begin{center}
\subfloat[][Evolution of the WCSoS and BCSoS for the k-Means- and Spherical k-Means-based universe matrices according to the number of centroids $k$, built upon FastText embeddings.]{\includegraphics[width=.48\textwidth]{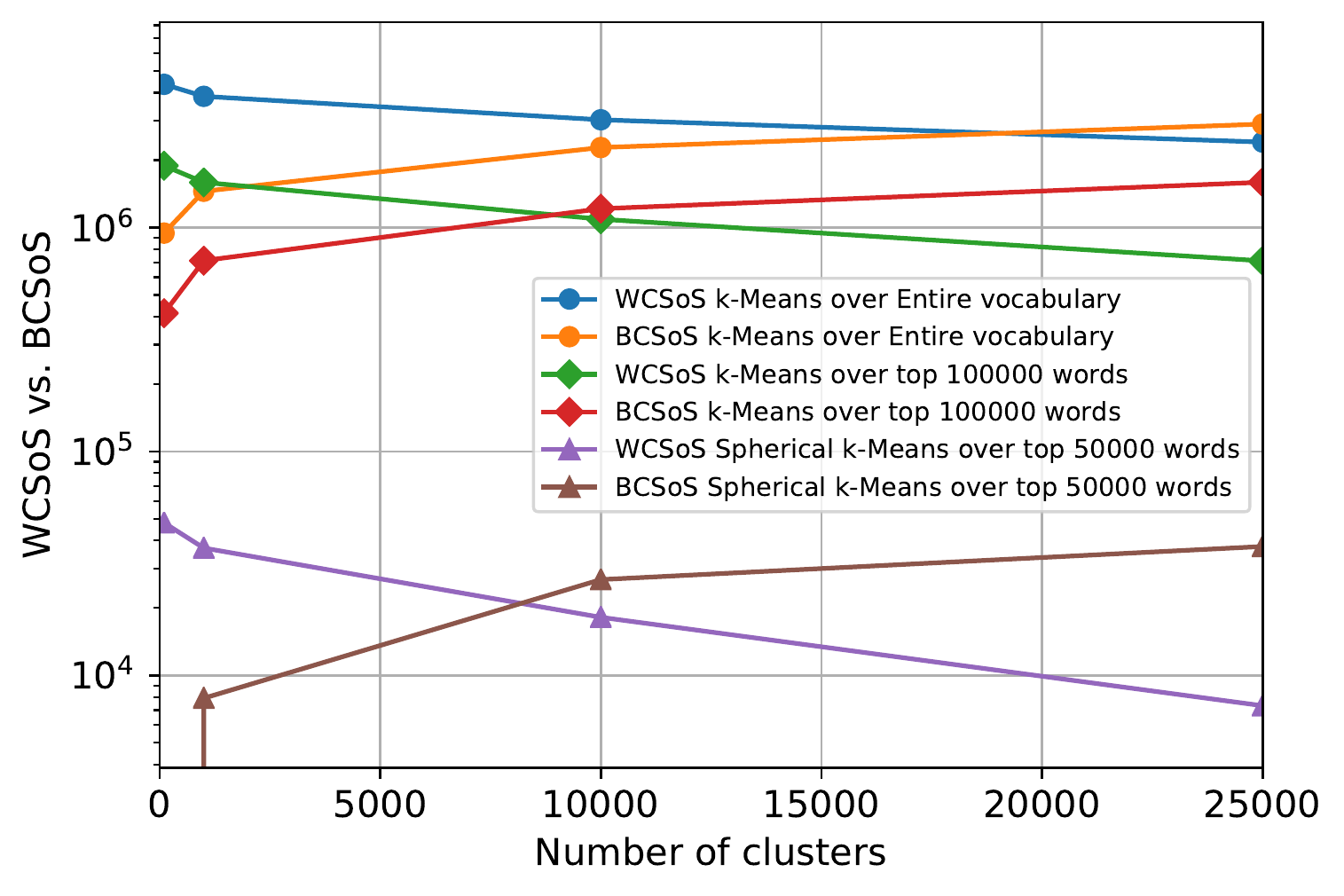}\label{fig:wbcsoskmeans}}\quad
\subfloat[][Explained variance of each dimension computed by PCA in order of components importance of all the considered word embedding vectors.]{\includegraphics[width=.48\textwidth]{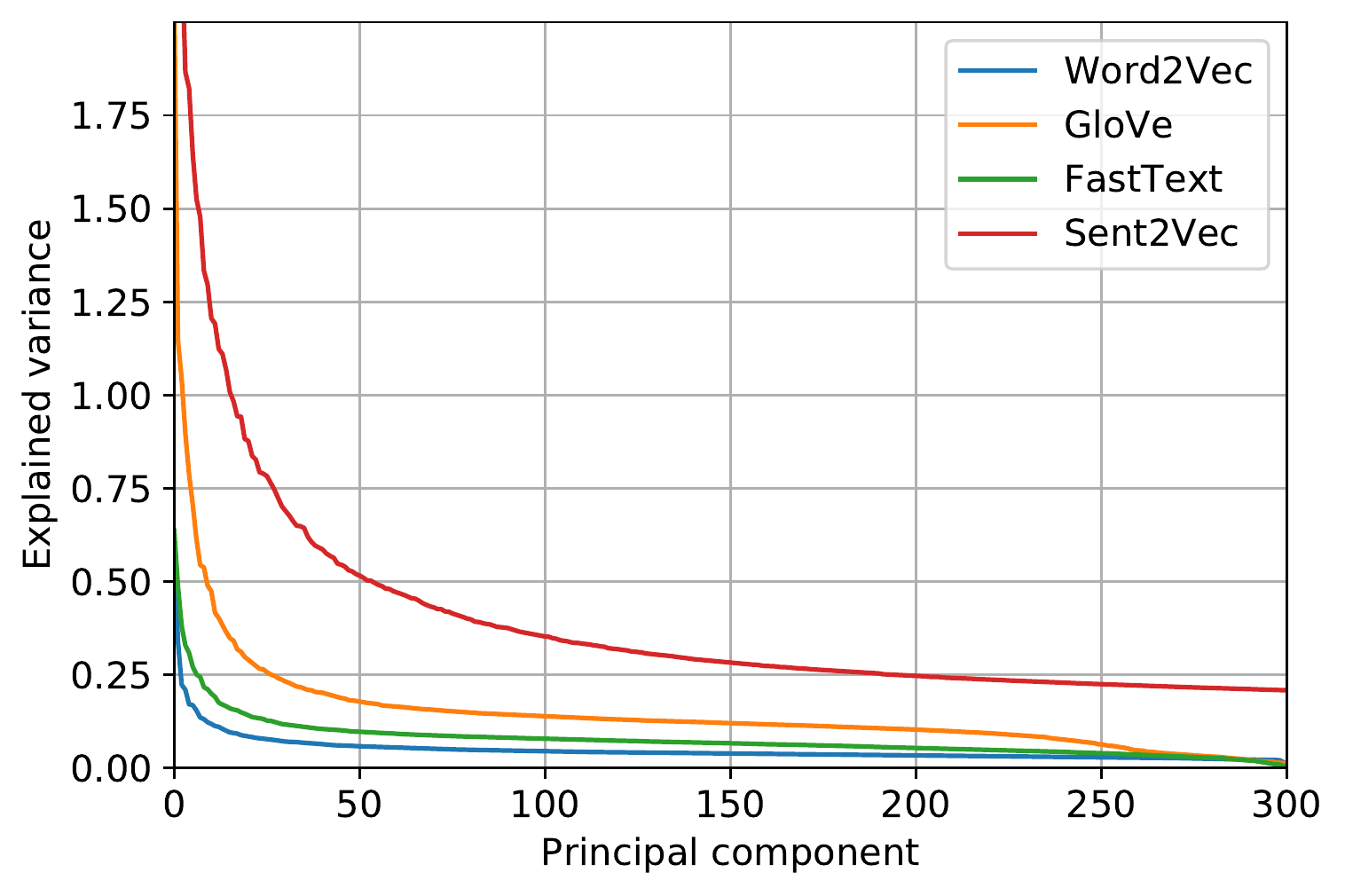}\label{fig:variancepca}}
\caption{Visualisation of the different analyses for the construction of the universe matrices.}
\label{fig:univermatrixanalysis}
\end{center}
\vspace{-5mm}
\end{figure}

For the sake of readability, in in Figure~\ref{fig:wbcsoskmeans}, we report only the clustering analysis of the universe matrices extracted from the FastText embeddings, as they turned out to provide the final best results.
The analysis of clustering results is conducted on the WCSoS and BCSoS for the different values of the number of clusters $k$.
In fact, from these graphs we can see that for all k-Means approaches the \emph{knee-elbow} point is positioned almost in the same spot, however the WCSoS and BCSoS are particularly high for the simple k-Means clusterings when compared to the Spherical one.
This might seem good in terms of BCSoS, but isn't for the WCSoS, which is preferable to be low.
Similarly we have the inverse situation when considering Spherical k-Means against the standard one, where the BCSoS is too low.

This behaviour highlights how none of the proposed clustering approaches behaves better than another when considered only in terms of clustering performances; however Spherical k-Means leads to a more promising universe matrix, among the clustering techniques we analysed (see Appendix~\hyperref[sec:extendedresults]{A}).

Interesting observations about PCA, instead, are reported in Figure~\ref{fig:variancepca}.
Here we show the evolution of the explained variance with the number of retained components (the graph is cut at $300$ components because all the word embeddings matrices except Sent2Vec have $300$ dimensions and because all of them reach and pass the ``elbow'' within that point).
Word2Vec and FastText are the ``fastest'' to converge, they retain most of the variance by the 50-th component, while Sent2Vec is the slowest, this is most probably due to the fact that Sent2Vec is built with more than double the dimensions of other embedding algorithms, hence the vectors are more scattered in the hyperspace.

As premised we compare our results with three baseline models and other sentence embedding approaches, all reported in Table~\ref{tab:resultscomparison}.
The first group of scores is from the baselines, the second one is from other sentence embedding models and, finally, the last group is from our SFBoW model.
Additionally, the best values in each column are highlighted in bold, while the second ones are underlined.


Sentence-BERT \cite{reimers2019sentencebert} outperforms all the other models in every task.
This was predictable as in relies on a much bigger feature extraction model and was trained on a much bigger corpus. 
However this model requires a considerably higher computational effort without an equally consistent difference in performances.
In fact BERT alone requires more than 100 million parameters just for its base version (and above 300 millions for the large one), hence taking a lot of (memory) space, not to mention the amount of time necessary for the self-supervised training and the fine-tuning.

Differently, non-parametric models (like those from \newcite{arora2019simple} and \newcite{zhelezniak2019dont} as well as SFBoW) or shallow parametric ones (Sent2Vec from \newcite{Pagliardini_2018}) require way less parameters -- just those for the embedding matrix $\left|\mathcal{V}\right| \times {d}$ -- and little, if any, training.
Nevertheless, given the low computational demand, they're all able to reach significant performances.

When compared to other models, SFBoW results are interesting: either considering majority of tasks with higher Spearman's $\rho$ rank or higher average score, it outperforms all the baselines, as well as SIF weighting and Sent2Vec.
Finally we see as our model performs closely to its predecessor, especially considering the weighted average of the single tasks results.
SFBoW bests out DynaMax in STS 2014 and gets almost the same results in STS 2012 (the difference of the two $\rho$ is $0.01$), which are the first two corpora in terms of samples; however the significant difference in STS 2013 goes in favour of DynaMax.

About the comparison against DynaMax, it is worth to underline a few additional points: first of all, in both cases fuzzy Jaccard similarity correlates better with human judgment as measure of sentence similarity.
Secondly both models manage to achieve better results when using FastText word embedding, possibly underling that they lend better than other models at sentence level combination; this is also shown by the baseline performances.
Finally we remind that DynaMax, differently from SFBoW, generates embeddings with variable size, which results in a limited applicability. 

\section{Conclusions}
\label{sec:conclusionsandfutureworks}
With this work we have proposed SFBoW, a refinement of the FBoW model for sentence embedding.
As its predecessor, our model is non-parametric (i.e. it doesn't require any training) and relies on fuzzy sets theory.
In fact, it leverages the fuzzy Jaccard index to compute semantic similarity from the sentence embeddings (differently from most of the other models, which rely on cosine similarity).

Even if SFBoW doesn't achieve state-of-the-art performances on the considered STS benchmark, our solution performs comparably to its predecessor, while enabling the possibility of re-usable embeddings as their dimension is fixed.
Moreover, as can be seen from the results, it outperforms in the majority of tasks all the other compared models except Sentence-BERT, without the need of specific training or fine tuning on sentence similarity corpora and still being as lightweight as possible.

As a future work, we plan to further investigate clustering techniques and, in general, other methodologies for computing the universe matrix.


\newpage
\bibliographystyle{coling}
\bibliography{coling2020}

\begin{thebibliography}{}

\bibitem[\protect\citename{Agirre \bgroup et al.\egroup }2012]{sts12}
Eneko Agirre, Daniel Cer, Mona Diab, and Aitor Gonzalez-Agirre.
\newblock 2012.
\newblock {S}em{E}val-2012 task 6: A pilot on semantic textual similarity.
\newblock In {\em *{SEM} 2012: The First Joint Conference on Lexical and
  Computational Semantics {--} Volume 1: Proceedings of the main conference and
  the shared task, and Volume 2: Proceedings of the Sixth International
  Workshop on Semantic Evaluation ({S}em{E}val 2012)}, pages 385--393,
  Montr{\'e}al, Canada, 7-8 June. Association for Computational Linguistics.

\bibitem[\protect\citename{Agirre \bgroup et al.\egroup }2013]{sts13}
Eneko Agirre, Daniel Cer, Mona Diab, Aitor Gonzalez-Agirre, and Weiwei Guo.
\newblock 2013.
\newblock {SEM} 2013 shared task: Semantic textual similarity.
\newblock In {\em Second Joint Conference on Lexical and Computational
  Semantics ({SEM}), Volume 1: Proceedings of the Main Conference and the
  Shared Task: Semantic Textual Similarity}, pages 32--43, Atlanta, Georgia,
  USA, June. Association for Computational Linguistics.

\bibitem[\protect\citename{Agirre \bgroup et al.\egroup }2014]{sts14}
Eneko Agirre, Carmen Banea, Claire Cardie, Daniel Cer, Mona Diab, Aitor
  Gonzalez-Agirre, Weiwei Guo, Rada Mihalcea, German Rigau, and Janyce Wiebe.
\newblock 2014.
\newblock {S}em{E}val-2014 task 10: Multilingual semantic textual similarity.
\newblock In {\em Proceedings of the 8th International Workshop on Semantic
  Evaluation ({S}em{E}val 2014)}, pages 81--91, Dublin, Ireland, August.
  Association for Computational Linguistics.

\bibitem[\protect\citename{Agirre \bgroup et al.\egroup }2015]{sts15}
Eneko Agirre, Carmen Banea, Claire Cardie, Daniel Cer, Mona Diab, Aitor
  Gonzalez-Agirre, Weiwei Guo, I{\~n}igo Lopez-Gazpio, Montse Maritxalar, Rada
  Mihalcea, German Rigau, Larraitz Uria, and Janyce Wiebe.
\newblock 2015.
\newblock {S}em{E}val-2015 task 2: Semantic textual similarity, {E}nglish,
  {S}panish and pilot on interpretability.
\newblock In {\em Proceedings of the 9th International Workshop on Semantic
  Evaluation ({S}em{E}val 2015)}, pages 252--263, Denver, Colorado, June.
  Association for Computational Linguistics.

\bibitem[\protect\citename{Agirre \bgroup et al.\egroup }2016]{sts16}
Eneko Agirre, Carmen Banea, Daniel Cer, Mona Diab, Aitor Gonzalez-Agirre, Rada
  Mihalcea, German Rigau, and Janyce Wiebe.
\newblock 2016.
\newblock {S}em{E}val-2016 task 1: Semantic textual similarity, monolingual and
  cross-lingual evaluation.
\newblock In {\em Proceedings of the 10th International Workshop on Semantic
  Evaluation ({S}em{E}val-2016)}, pages 497--511, San Diego, California, June.
  Association for Computational Linguistics.

\bibitem[\protect\citename{Arora \bgroup et al.\egroup }2019]{arora2019simple}
Sanjeev Arora, Yingyu Liang, and Tengyu Ma.
\newblock 2019.
\newblock A simple but tough-to-beat baseline for sentence embeddings.
\newblock In {\em 5th International Conference on Learning Representations,
  ICLR 2017}.

\bibitem[\protect\citename{Baroni \bgroup et al.\egroup
  }2014]{baroni-etal-2014-dont}
Marco Baroni, Georgiana Dinu, and Germ{\'a}n Kruszewski.
\newblock 2014.
\newblock Don{'}t count, predict! a systematic comparison of context-counting
  vs. context-predicting semantic vectors.
\newblock In {\em Proceedings of the 52nd Annual Meeting of the Association for
  Computational Linguistics (Volume 1: Long Papers)}, pages 238--247,
  Baltimore, Maryland, June. Association for Computational Linguistics.

\bibitem[\protect\citename{Bengio \bgroup et al.\egroup
  }2003]{bengio2003neural}
Yoshua Bengio, R{\'e}jean Ducharme, Pascal Vincent, and Christian Jauvin.
\newblock 2003.
\newblock A neural probabilistic language model.
\newblock {\em Journal of machine learning research}, 3(Feb):1137--1155.

\bibitem[\protect\citename{Bengio \bgroup et al.\egroup }2013]{6472238}
Yoshua Bengio, Aaron Courville, and Pascal Vincent.
\newblock 2013.
\newblock Representation learning: A review and new perspectives.
\newblock {\em IEEE Transactions on Pattern Analysis and Machine Intelligence},
  35(8):1798--1828, Aug.

\bibitem[\protect\citename{Bojanowski \bgroup et al.\egroup }2017]{fasttext}
Piotr Bojanowski, Edouard Grave, Armand Joulin, and Tomas Mikolov.
\newblock 2017.
\newblock Enriching word vectors with subword information.
\newblock {\em Transactions of the Association for Computational Linguistics},
  5:135--146.

\bibitem[\protect\citename{Bowman \bgroup et al.\egroup }2015]{snli}
Samuel~R. Bowman, Gabor Angeli, Christopher Potts, and Christopher~D. Manning.
\newblock 2015.
\newblock A large annotated corpus for learning natural language inference.
\newblock In {\em Proceedings of the 2015 Conference on Empirical Methods in
  Natural Language Processing}, pages 632--642, Lisbon, Portugal, September.
  Association for Computational Linguistics.

\bibitem[\protect\citename{Collobert and Weston}2007]{collobert2007fast}
Ronan Collobert and Jason Weston.
\newblock 2007.
\newblock Fast semantic extraction using a novel neural network architecture.
\newblock In {\em Proceedings of the 45th Annual Meeting of the Association of
  Computational Linguistics}, pages 560--567.

\bibitem[\protect\citename{Collobert and Weston}2008]{collobert2008unified}
Ronan Collobert and Jason Weston.
\newblock 2008.
\newblock A unified architecture for natural language processing: Deep neural
  networks with multitask learning.
\newblock In {\em Proceedings of the 25th international conference on Machine
  learning}, pages 160--167.

\bibitem[\protect\citename{Collobert \bgroup et al.\egroup
  }2011]{collobert2011nlpscratch}
Ronan Collobert, Jason Weston, Leon Bottou, Michael Karlen, Koray Kavukcuoglu,
  and Pavel Kuksa.
\newblock 2011.
\newblock Natural language processing (almost) from scratch.
\newblock {\em Computing Research Repository - CORR}, 12, 03.

\bibitem[\protect\citename{Deerwester \bgroup et al.\egroup
  }1989]{deerwester1989computer}
Scott~C. Deerwester, Susan~T. Dumais, George~W. Furnas, Richard~A. Harshman,
  Thomas~K. Landauer, Karen~E. Lochbaum, and Lynn~A. Streeter.
\newblock 1989.
\newblock Computer information retrieval using latent semantic structure,
  June~13.

\bibitem[\protect\citename{Devlin \bgroup et al.\egroup }2018]{devlin2018bert}
Jacob Devlin, Ming-Wei Chang, Kenton Lee, and Kristina Toutanova.
\newblock 2018.
\newblock Bert: Pre-training of deep bidirectional transformers for language
  understanding.

\bibitem[\protect\citename{Ester \bgroup et al.\egroup
  }1996]{Ester96adensity-based}
Martin Ester, Hans-Peter Kriegel, J{\"o}rg Sander, Xiaowei Xu, et~al.
\newblock 1996.
\newblock A density-based algorithm for discovering clusters in large spatial
  databases with noise.
\newblock In {\em Kdd}. AAAI Press.

\bibitem[\protect\citename{Gupta \bgroup et al.\egroup }2019]{sent2vec_word}
Prakhar Gupta, Matteo Pagliardini, and Martin Jaggi.
\newblock 2019.
\newblock Better word embeddings by disentangling contextual n-gram
  information.
\newblock {\em arXiv preprint arXiv:1904.05033}.

\bibitem[\protect\citename{He and Yih}2014]{he2014deep}
Xiaodong He and Scott Wen-tau Yih.
\newblock 2014.
\newblock Deep learning and continuous representations for language processing
  (tutorial for slt-2014), December.

\bibitem[\protect\citename{Hornik \bgroup et al.\egroup
  }2012]{hornik2012spherical}
Kurt Hornik, Ingo Feinerer, Martin Kober, and Christian Buchta.
\newblock 2012.
\newblock Spherical k-means clustering.
\newblock {\em Journal of Statistical Software}, 50:1--22, 09.

\bibitem[\protect\citename{Kiros \bgroup et al.\egroup }2015]{kiros2015skip}
Ryan Kiros, Yukun Zhu, Russ~R Salakhutdinov, Richard Zemel, Raquel Urtasun,
  Antonio Torralba, and Sanja Fidler.
\newblock 2015.
\newblock Skip-thought vectors.
\newblock In {\em Advances in neural information processing systems}, pages
  3294--3302.

\bibitem[\protect\citename{MacQueen}1967]{macqueen1967}
James MacQueen.
\newblock 1967.
\newblock Some methods for classification and analysis of multivariate
  observations.
\newblock In {\em Proceedings of the Fifth Berkeley Symposium on Mathematical
  Statistics and Probability, Volume 1: Statistics}, pages 281--297, Berkeley,
  Calif. University of California Press.

\bibitem[\protect\citename{Mikolov \bgroup et al.\egroup
  }2013a]{mikolov2013efficient}
Tomas Mikolov, Kai Chen, Greg Corrado, and Jeffrey Dean.
\newblock 2013a.
\newblock Efficient estimation of word representations in vector space.

\bibitem[\protect\citename{Mikolov \bgroup et al.\egroup
  }2013b]{mikolov2013distributed}
Tomas Mikolov, Ilya Sutskever, Kai Chen, Greg~S Corrado, and Jeff Dean.
\newblock 2013b.
\newblock Distributed representations of words and phrases and their
  compositionality.
\newblock In {\em Advances in neural information processing systems}, pages
  3111--3119.

\bibitem[\protect\citename{Osgood \bgroup et al.\egroup
  }1958]{osgood1957measurement}
Charles~Egerton Osgood, George~J Suci, and Percy~H. Tannenbaum.
\newblock 1958.
\newblock The measurement of meaning.
\newblock {\em American Journal of Sociology}, 63(5):550--551.

\bibitem[\protect\citename{Pagliardini \bgroup et al.\egroup
  }2018]{Pagliardini_2018}
Matteo Pagliardini, Prakhar Gupta, and Martin Jaggi.
\newblock 2018.
\newblock Unsupervised learning of sentence embeddings using compositional
  n-gram features.
\newblock {\em Proceedings of the 2018 Conference of the North American Chapter
  of the Association for Computational Linguistics: Human Language
  Technologies, Volume 1 (Long Papers)}.

\bibitem[\protect\citename{Pennington \bgroup et al.\egroup
  }2014]{pennington2014glove}
Jeffrey Pennington, Richard Socher, and Christopher~D. Manning.
\newblock 2014.
\newblock Glove: Global vectors for word representation.
\newblock In {\em Empirical Methods in Natural Language Processing (EMNLP)},
  pages 1532--1543.

\bibitem[\protect\citename{Radford \bgroup et al.\egroup
  }2018]{radford2018improving}
Alec Radford, Karthik Narasimhan, Tim Salimans, and Ilya Sutskever.
\newblock 2018.
\newblock Improving language understanding by generative pre-training.
\newblock {\em URL https://s3-us-west-2. amazonaws.
  com/openai-assets/researchcovers/languageunsupervised/language understanding
  paper. pdf}.

\bibitem[\protect\citename{Radford \bgroup et al.\egroup
  }2019]{radford2019language}
Alec Radford, Jeffrey Wu, Rewon Child, David Luan, Dario Amodei, and Ilya
  Sutskever.
\newblock 2019.
\newblock Language models are unsupervised multitask learners.
\newblock {\em OpenAI Blog}, 1(8):9.

\bibitem[\protect\citename{Reimers and Gurevych}2019]{reimers2019sentencebert}
Nils Reimers and Iryna Gurevych.
\newblock 2019.
\newblock Sentence-bert: Sentence embeddings using siamese bert-networks.

\bibitem[\protect\citename{Reimers \bgroup et al.\egroup
  }2016]{reimers-etal-2016-task}
Nils Reimers, Philip Beyer, and Iryna Gurevych.
\newblock 2016.
\newblock Task-oriented intrinsic evaluation of semantic textual similarity.
\newblock In {\em Proceedings of {COLING} 2016, the 26th International
  Conference on Computational Linguistics: Technical Papers}, pages 87--96,
  Osaka, Japan, December. The COLING 2016 Organizing Committee.

\bibitem[\protect\citename{Salton}1971]{salton1971smart}
Gerard Salton.
\newblock 1971.
\newblock {\em The SMART retrieval system: experiments in automatic document
  processing}.
\newblock Prentice-Hall series in automatic computation. Prentice-Hall.

\bibitem[\protect\citename{Sordoni \bgroup et al.\egroup
  }2015]{sordoni-etal-2015-neural}
Alessandro Sordoni, Michel Galley, Michael Auli, Chris Brockett, Yangfeng Ji,
  Margaret Mitchell, Jian-Yun Nie, Jianfeng Gao, and Bill Dolan.
\newblock 2015.
\newblock A neural network approach to context-sensitive generation of
  conversational responses.
\newblock In {\em Proceedings of the 2015 Conference of the North {A}merican
  Chapter of the Association for Computational Linguistics: Human Language
  Technologies}, pages 196--205, Denver, Colorado, May{--}June. Association for
  Computational Linguistics.

\bibitem[\protect\citename{Sparck~Jones}1986]{10.5555/22908}
Karen Sparck~Jones.
\newblock 1986.
\newblock {\em Synonymy and Semantic Classification}.
\newblock Edinburgh University Press, GBR.

\bibitem[\protect\citename{Sutskever \bgroup et al.\egroup
  }2014]{sutskever2014sequence}
Ilya Sutskever, Oriol Vinyals, and Quoc~V Le.
\newblock 2014.
\newblock Sequence to sequence learning with neural networks.
\newblock In {\em Advances in neural information processing systems}, pages
  3104--3112.

\bibitem[\protect\citename{Vaswani \bgroup et al.\egroup
  }2017]{vaswani2017attention}
Ashish Vaswani, Noam Shazeer, Niki Parmar, Jakob Uszkoreit, Llion Jones,
  Aidan~N Gomez, {\L}ukasz Kaiser, and Illia Polosukhin.
\newblock 2017.
\newblock Attention is all you need.
\newblock In {\em Advances in neural information processing systems}, pages
  5998--6008.

\bibitem[\protect\citename{Wang \bgroup et al.\egroup }2019]{wang2019superglue}
Alex Wang, Yada Pruksachatkun, Nikita Nangia, Amanpreet Singh, Julian Michael,
  Felix Hill, Omer Levy, and Samuel Bowman.
\newblock 2019.
\newblock Super{GLUE}: A stickier benchmark for general-purpose language
  understanding systems.
\newblock {\em arXiv preprint 1905.00537}.

\bibitem[\protect\citename{Williams \bgroup et al.\egroup }2018]{snli2}
Adina Williams, Nikita Nangia, and Samuel Bowman.
\newblock 2018.
\newblock A broad-coverage challenge corpus for sentence understanding through
  inference.
\newblock In {\em Proceedings of the 2018 Conference of the North {A}merican
  Chapter of the Association for Computational Linguistics: Human Language
  Technologies, Volume 1 (Long Papers)}, pages 1112--1122, New Orleans,
  Louisiana, June. Association for Computational Linguistics.

\bibitem[\protect\citename{Yang \bgroup et al.\egroup }2019]{yang2019parameter}
Ziyi Yang, Chenguang Zhu, and Weizhu Chen.
\newblock 2019.
\newblock Parameter-free sentence embedding via orthogonal basis.
\newblock In {\em Proceedings of the 2019 Conference on Empirical Methods in
  Natural Language Processing and the 9th International Joint Conference on
  Natural Language Processing (EMNLP-IJCNLP)}, pages 638--648.

\bibitem[\protect\citename{Zadeh}1965]{zadeh}
Lotfi~A. Zadeh.
\newblock 1965.
\newblock Fuzzy sets.
\newblock {\em Information and Control}, 8(3):338 -- 353.

\bibitem[\protect\citename{Zhao and Mao}2018]{zhao_e_mao}
Rui Zhao and Kezhi Mao.
\newblock 2018.
\newblock Fuzzy bag-of-words model for document representation.
\newblock {\em IEEE Transactions on Fuzzy Systems}, 26(2):794--804, April.

\bibitem[\protect\citename{Zhelezniak \bgroup et al.\egroup
  }2019]{zhelezniak2019dont}
Vitalii Zhelezniak, Aleksandar Savkov, April Shen, Francesco Moramarco, Jack
  Flann, and Nils~Y. Hammerla.
\newblock 2019.
\newblock Don't settle for average, go for the max: Fuzzy sets and max-pooled
  word vectors.

\end{thebibliography}

\newpage
\appendix

\section*{Appendix A. Extended results}
\label{sec:extendedresults}
In this appendix we report the complete results of our experiments.
Tables~\ref{tab:globalsummary} and \ref{tab:univsummary}, contain the average score on each task fixing, respectively, the word embedding matrix and the universe matrix construction approach.
These first two tables are provided to ease the understanding of numeric results.

Additionally, Tables~\ref{tab:w2vsummary}, \ref{tab:gvsummary}, \ref{tab:ftsummary} and \ref{tab:s2vsummary} contain more detailed results of the approaches we used to build the universe matrix using, respectively, the following word embedding matrices: Word2Vec, GloVe, FastText and Sent2Vec.
In these tables is possible to find, for each word embedding used to build the universe matrix, the following contents:
\begin{itemize}
\setlength\itemsep{0em}
\item complete scores over each benchmark;
\item summary scores of different approaches using the same word embedding matrix (rightmost column);
\item the best score on each benchmark and on average of each approach;
\item summary of all the approaches using the same word embedding matrix (bottom row).
\end{itemize}

In all the following tables, summary statistics are expressed as $\textit{average score} \pm \textit{standard deviation}$.

The first thing that comes to the eye in these table, is the common behaviour on tasks shared by all models independently of the word embedding matrix (see Table~\ref{tab:globalsummary}) or the universe matrix construction approach (see Table~\ref{tab:univsummary}).
In fact, they all struggle more with the 2013 edition of the STS (in each row it has always the lowest Spearman index) while they perform the best with the 2015 and 2016 editions.

Another interesting point is in the standard deviation of the summary statistics: while the value appear to be high for all the models, when averaging among the different tasks (as shown in the rightmost columns of Tables~\ref{tab:w2vsummary}, \ref{tab:gvsummary}, \ref{tab:ftsummary} and \ref{tab:s2vsummary}), this doesn't hold when averaging over the same task with different models (downmost row of the aforementioned Tables).
Models based on Sent2Vec have a much lower standard deviation in this case, as can be seen in Table~\ref{tab:s2vsummary}, while for other word embedding matrices the standard deviation is similar to that computed among the tasks.

About Sent2Vec it is curious how far it gets from the best results in our experiments, especially considering that differently from the other embedding models we used, it was conceived specifically for sentence embedding purposes (see Table~\ref{tab:s2vsummary}).
It is also worth to mention that, even though it is a sentence embedding algorithm, Sent2Vec performs better using a universe matrix obtained through Spherical k-Means than the Identity matrix (as we were expecting), as reported in Table~\ref{tab:s2vsummary}.
This is the only case where a SFBoW model has better results in single tasks or in average with a clustering built universe matrix instead of Identity- or PCA-built, as happened with the other embedding matrices (see Tables~\ref{tab:w2vsummary}, \ref{tab:gvsummary} and \ref{tab:ftsummary}).

Finally we underline a surprising behaviour of the FastText based models.
Even though SFBoW built upon FastText reaches the best results, on average, considering all the approaches for the universe matrix, they perform worse than other models.
In fact, when used with clustering, results are way lower than Identity and PCA; this can be seen from Table~\ref{tab:ftsummary}.

\begin{table}[ht]
\caption{Summarised results (measured as Spearman's correlation with human judgment) of the SFBoW model on the STS tasks using various word embedding matrices as source. Values are computed averaging the results for different universe matrices. Bold values represent the best score in each column.}
\begin{center}
\begin{tabularx}{\textwidth}{B | ccccc}
\toprule
\multirow{3}{=}{Word embedding matrix} & \multicolumn{5}{c}{Results (Spearman's $\rho$)} \\ \cline{2-6}
& \multicolumn{5}{c}{STS} \\ \cline{2-6}
& '12 & '13 & '14 & '15 & '16 \\ 
\midrule
Word2Vec & $52.82\pm3.54$ & $\mathbf{44.18}\pm4.83$ & $\mathbf{59.47}\pm4.32$ & $64.97\pm5.09$ & $\mathbf{64.52}\pm4.73$ \\ 
GloVe & $53.77\pm4.69$ & $43.84\pm5.61$ & $55.36\pm7.30$ & $62.38\pm7.65$ & $63.56\pm6.20$ \\ 
FastText & $\mathbf{54.99}\pm4.88$ & $43.65\pm4.03$ & $54.36\pm5.63$ & $63.75\pm6.65$ & $64.40\pm6.22$ \\ 
Sent2Vec & $53.12\pm1.34$ & $41.58\pm2.32$ & $59.24\pm2.78$ & $\mathbf{64.99}\pm2.86$ & $62.64\pm2.24$ \\ 
\bottomrule
\end{tabularx}
\end{center}
\label{tab:globalsummary}
\end{table}%

\begin{sidewaystable}[htp]
\caption{Summarised results (measured as Spearman's correlation with human judgment) of the SFBoW model on the STS tasks using various universe matrices. Values are computed averaging the results for different word embedding matrices. Bold values represent the best score in each column.}
\begin{center}
\begin{tabular}{l | l | l | ccccc}
\toprule
\multicolumn{3}{l |}{} & \multicolumn{5}{c}{Results (Spearman's $\rho$)} \\ \cline{4-8}
\multicolumn{3}{l |}{Universe matrix} &\multicolumn{5}{c}{STS} \\ \cline{4-8}
\multicolumn{3}{l |}{} & '12 & '13 & '14 & '15 & '16 \\
\midrule
\multicolumn{8}{l}{Clustering} \\
\midrule
Algorithm & Source & $k$ &  \multicolumn{5}{c}{} \\ \cline{1-8}
\multirow{8}{*}{k-Means} & \multirow{4}{*}{English vocabulary} & 100 & $47.24\pm2.85$ & $36.29\pm1.05$ & $47.88\pm5.20$ & $53.81\pm5.17$ & $55.26\pm2.06$  \\
& & 1000 & $50.10\pm2.01$ & $39.70\pm1.00$ & $52.04\pm4.69$ & $58.49\pm4.48$ & $59.48\pm2.05$ \\
& & 10000 & $52.04\pm2.26$ & $42.66\pm2.00$ & $55.26\pm4.98$ & $61.14\pm5.13$ & $62.16\pm2.16$ \\
& & 25000 & $52.34\pm3.29$ & $42.91\pm2.56$ & $55.75\pm6.21$ & $61.56\pm6.34$ & $62.60\pm3.05$ \\ \cline{2-8}
& \multirow{4}{*}{Frequent words (top 100000)} & 100 & $50.47\pm2.70$ & $38.34\pm2.70$ & $52.39\pm2.45$ & $58.16\pm2.90$ & $58.15\pm2.60$ \\
& & 1000 & $52.89\pm3.56$ & $42.67\pm3.35$ & $56.38\pm2.97$ & $62.42\pm3.64$ & $62.14\pm3.58$ \\
& & 10000 & $54.64\pm3.03$ & $45.34\pm3.12$ & $59.48\pm2.01$ & $65.37\pm2.72$ & $66.42\pm3.05$  \\
& & 25000 & $55.33\pm2.47$ & $46.18\pm2.83$ & $60.43\pm1.93$ & $66.41\pm2.39$ & $66.33\pm2.81$ \\ \cline{1-8}
\multirow{4}{*}{Spherical k-Means} & \multirow{4}{*}{Frequent words (top 50000)} & 100 &  $51.82\pm0.99$ & $39.23\pm3.13$ & $53.83\pm3.93$ & $60.77\pm2.52$ & $59.92\pm3.32$ \\
& & 1000 & $54.32\pm1.48$ & $43.33\pm3.26$ & $57.60\pm3.49$ & $64.89\pm2.56$ & $63.52\pm3.07$ \\
& & 10000 & $55.95\pm1.69$ & $46.22\pm2.60$ & $60.67\pm2.48$ & $67.88\pm1.92$ & $66.36\pm2.18$ \\
& & 25000 & $56.35\pm1.52$ & $46.01\pm2.53$ & $60.92\pm2.37$ & $68.20\pm1.73$ & $66.48\pm2.02$ \\ 
\midrule
\multicolumn{8}{l}{Identity matrix} \\
\midrule
\multicolumn{3}{c |}{} & $56.91\pm3.35$ & $\mathbf{49.45}\pm3.12$ & $\mathbf{64.56}\pm1.90$ & $\mathbf{70.59}\pm1.58$ & $69.33\pm3.63$ \\
\midrule
\multicolumn{8}{l}{Multivariate analysis} \\ 
\midrule
Algorithm& \multicolumn{2}{l |}{Source} &  \multicolumn{5}{c}{} \\ \cline{1-8}
\multirow{2}{*}{PCA} &  \multicolumn{2}{l |}{English vocabulary} &  $\mathbf{58.06}\pm3.36$ & $48.83\pm3.34$ & $64.35\pm1.69$ & $\mathbf{70.59}\pm1.77$ & $\mathbf{69.96}\pm3.99$ \\
& \multicolumn{2}{l |}{Frequent words (top 100000)} &  $57.00\pm2.62$ & $48.45\pm3.08$ & $64.20\pm1.60$ & $69.81\pm1.68$ & $69.64\pm3.52$ \\
\bottomrule
\end{tabular}
\end{center}
\label{tab:univsummary}
\end{sidewaystable}%

\begin{sidewaystable}[htp]
\caption{Complete experimental results (measured as Spearman's correlation with human judgment) of the SFBoW model on the STS tasks using a Word2Vec matrix as word embedding source. On the rightmost columns are reported an unweighted and a weighted per-model summary over the tasks, while on the bottom row is reported a per-task summary over the models, finally on the bottom-right corner are the unweighted and weighted per-word embedding matrix statistics.  Bold values represent the highest score in the corresponding column.}
\begin{center}
\begin{tabularx}{\textwidth}{B | B | S | lllll | c | c}
\toprule
\multicolumn{3}{l |}{} & \multicolumn{7}{c}{Results (Spearman's $\rho$)} \\ \cline{4-10}
\multicolumn{3}{l |}{Universe matrix} &\multicolumn{5}{c |}{STS} & \multirow{2}{*}{Average} & \multirow{2}{*}{Weighted average} \\ \cline{4-8}
\multicolumn{3}{l |}{} & '12 & '13 & '14 & '15 & '16 & & \\
\midrule
\multicolumn{10}{l}{Clustering} \\
\midrule
Algorithm & Source & $k$ &  \multicolumn{7}{c}{} \\ \cline{1-10}
\multirow{8}{=}{k-Means} & \multirow{4}{=}{English vocabulary} & 100 & $48.75$ & $35.10$ & $51.32$ & $56.58$ & $56.01$ & $49.55\pm7.79$ & $49.46\pm6.65$ \\
& & 1000 & $52.10$ & $41.29$ & $57.15$ & $62.78$ & $61.91$ & $55.05\pm7.86$ & $54.58\pm6.83$ \\
& & 10000 & $54.05$ & $45.63$ & $60.90$ & $66.51$ & $65.13$ & $58.44\pm7.74$ & $57.76\pm6.90$ \\
& & 25000 & $54.53$ & $45.79$ & $61.24$ & $66.86$ & $65.14$ & $58.71\pm7.73$ & $58.10\pm6.89$ \\ \cline{2-10}
& \multirow{4}{=}{Frequent words (top 100000)} & 100 & $46.51$ & $35.11$ & $51.47$ & $54.47$ & $55.93$ & $48.70\pm7.52$ & $48.32\pm6.38$ \\
& & 1000 & $47.41$ & $39.14$ & $53.62$ & $57.78$ & $58.24$ & $51.24\pm7.19$ & $50.56\pm6.27$ \\
& & 10000 & $50.17$ & $43.39$ & $58.03$ & $62.47$ & $63.16$ & $55.44\pm7.60$ & $54.48\pm6.75$ \\
& & 25000 & $51.74$ & $44.63$ & $59.83$ & $64.55$ & $64.73$ & $57.10\pm7.82$ & $56.16\pm6.97$ \\ \cline{1-10}
\multirow{4}{=}{Spherical k-Means} & \multirow{4}{=}{Frequent words (top 50000)} & 100 & $51.40$ & $41.11$ & $58.33$ & $64.20$ & $63.65$ & $55.74\pm8.65$ & $55.01\pm7.58$ \\
& & 1000 & $53.50$ & $45.58$ & $61.19$ & $68.08$ & $66.86$ & $59.04\pm8.48$ & $58.07\pm7.59$ \\
& & 10000 & $54.49$ & $47.70$ & $62.84$ & $69.92$ & $67.85$ & $60.56\pm8.34$ & $59.55\pm7.62$ \\
& & 25000 & $55.18$ & $47.61$ & $63.04$ & $69.88$ & $67.77$ & $60.70\pm8.27$ & $59.81\pm7.49$ \\
\midrule
\multicolumn{10}{l}{Identity matrix} \\
\midrule
\multicolumn{3}{c |}{} & $54.69$ & $\mathbf{50.99}$ & $63.81$ & $69.80$ & $68.52$ & $61.56\pm7.49$ & $60.37\pm7.00$\\
\midrule
\multicolumn{10}{l}{Multivariate analysis} \\ 
\midrule
Algorithm& \multicolumn{2}{Z |}{Source} &  \multicolumn{7}{c}{} \\ \cline{1-10}
\multirow{2}{*}{PCA} &  \multicolumn{2}{Z |}{English vocabulary} & $\mathbf{59.78}$ & $50.26$ & $64.79$ & $\mathbf{70.61}$ & $\mathbf{71.81}$ & $\mathbf{63.45}\pm7.88$ & $\mathbf{62.84}\pm6.71$\\
& \multicolumn{2}{Z |}{Frequent words (top 100000)} & $57.97$ & $49.42$ & $\mathbf{64.88}$ & $70.10$ & $71.12$ & $62.70\pm8.11$ & $61.77\pm7.05$\\
\midrule
\multicolumn{10}{l}{} \\
\midrule
\multicolumn{3}{l |}{Average} & $52.82\pm3.54$ & $44.18\pm4.83$ & $59.47\pm4.32$ & $64.97\pm5.09$ & $64.52\pm4.73$ & & \\
\bottomrule
\end{tabularx}
\end{center}
\label{tab:w2vsummary}
\end{sidewaystable}

\begin{sidewaystable}[htp]
\caption{Complete experimental results (measured as Spearman's correlation with human judgment) of the SFBoW model on the STS tasks using a GloVe matrix as word embedding source. On the rightmost columns are reported an unweighted and a weighted per-model summary over the tasks, while on the bottom row is reported a per-task summary over the models, finally on the bottom-right corner are the unweighted and weighted per-word embedding matrix statistics. Bold values represent the highest score in the corresponding column.}
\begin{center}
\begin{tabularx}{\textwidth}{B | B | S | lllll | c | c}
\toprule
\multicolumn{3}{l |}{} & \multicolumn{7}{c}{Results (Spearman's $\rho$)} \\ \cline{4-10}
\multicolumn{3}{l |}{Universe matrix} &\multicolumn{5}{c |}{STS} & \multirow{2}{*}{Average} & \multirow{2}{*}{Weighted average} \\ \cline{4-8}
\multicolumn{3}{l |}{} & '12 & '13 & '14 & '15 & '16 & & \\
\midrule
\multicolumn{10}{l}{Clustering} \\
\midrule
Algorithm & Source & $k$ &  \multicolumn{7}{c}{} \\ \cline{1-10}
\multirow{8}{=}{k-Means} & \multirow{4}{=}{English vocabulary} & 100 & $44.94$ & $36.34$ & $43.07$ & $47.75$ & $53.50$ & $45.12\pm5.63$ & $44.44\pm4.31$ \\
& & 1000 & $47.33$ & $38.62$ & $46.72$ & $52.57$ & $57.24$ & $48.50\pm6.24$ & $47.69\pm4.89$ \\
& & 10000 & $48.23$ & $39.99$ & $48.38$ & $53.91$ & $59.17$ & $49.94\pm6.41$ & $49.00\pm4.98$ \\
& & 25000 & $46.71$ & $38.78$ & $46.43$ & $51.52$ & $57.71$ & $48.23\pm6.25$ & $47.27\pm4.76$ \\ \cline{2-10}
& \multirow{4}{=}{Frequent words (top 100000)} & 100 & $53.88$ & $42.42$ & $54.79$ & $61.83$ & $62.14$ & $55.01\pm7.17$ & $54.61\pm6.03$ \\
& & 1000 & $57.34$ & $48.07$ & $60.09$ & $67.34$ & $67.87$ & $60.14\pm7.28$ & $59.41\pm6.17$ \\
& & 10000 & $58.35$ & $50.21$ & $62.50$ & $69.51$ & $70.72$ & $62.26\pm7.55$ & $61.29\pm6.41$ \\
& & 25000 & $58.52$ & $50.88$ & $62.86$ & $70.06$ & $71.20$ & $62.70\pm7.53$ & $61.67\pm6.43$ \\ \cline{1-10}
\multirow{4}{=}{Spherical k-Means} & \multirow{4}{=}{Frequent words (top 50000)} & 100 & $51.57$ & $34.52$ & $47.52$ & $57.28$ & $54.84$ & $49.15\pm8.01$ & $49.46\pm7.00$ \\
& & 1000 & $53.53$ & $39.32$ & $51.85$ & $61.52$ & $59.13$ & $53.07\pm7.73$ & $53.03\pm6.69$ \\
& & 10000 & $55.58$ & $43.74$ & $56.50$ & $65.35$ & $63.61$ & $56.96\pm7.64$ & $56.59\pm6.56$ \\
& & 25000 & $56.10$ & $43.63$ & $56.91$ & $65.92$ & $63.90$ & $57.29\pm7.83$ & $56.99\pm6.74$ \\
\midrule
\multicolumn{10}{l}{Identity matrix} \\
\midrule
\multicolumn{3}{c |}{} & $\mathbf{58.82}$ & $\mathbf{51.55}$ & $\mathbf{64.74}$ & $71.03$ & $70.96$ & $\mathbf{63.42}\pm7.46$ & $\mathbf{62.50}\pm6.55$ \\
\midrule
\multicolumn{10}{l}{Multivariate analysis} \\ 
\midrule
Algorithm& \multicolumn{2}{Z |}{Source} &  \multicolumn{7}{c}{} \\ \cline{1-10}
\multirow{2}{*}{PCA} &  \multicolumn{2}{Z |}{English vocabulary} & $58.53$ & $50.59$ & $64.46$ & $\mathbf{71.18}$ & $\mathbf{71.04}$ & $63.16\pm7.84$ & $62.23\pm6.86$ \\
& \multicolumn{2}{Z |}{Frequent words (top 100000)} & $57.15$ & $48.91$ & $63.51$ & $68.89$ & $70.36$ & $61.76\pm7.93$ & $60.79\pm6.83$ \\
\midrule
\multicolumn{10}{l}{} \\
\midrule
\multicolumn{3}{l |}{Average} & $53.77\pm4.69$ & $43.84\pm5.61$ & $55.36\pm7.30$ & $62.38\pm7.65$ & $63.56\pm6.20$ & & \\
\bottomrule
\end{tabularx}
\end{center}
\label{tab:gvsummary}
\end{sidewaystable}

\begin{sidewaystable}[htp]
\caption{Complete experimental results (measured as Spearman's correlation with human judgment) of the SFBoW model on the STS tasks using a FastText matrix as word embedding source. On the rightmost columns are reported an unweighted and a weighted per-model summary over the tasks, while on the bottom row is reported a per-task summary over the models, finally on the bottom-right corner are the unweighted and weighted per-word embedding matrix statistics. Bold values represent the highest score in the corresponding column.}
\begin{center}
\begin{tabularx}{\textwidth}{B | B | S | lllll | c | c}
\toprule
\multicolumn{3}{l |}{} & \multicolumn{7}{c}{Results (Spearman's $\rho$)} \\ \cline{4-10}
\multicolumn{3}{l |}{Universe matrix} &\multicolumn{5}{c |}{STS} & \multirow{2}{*}{Average} & \multirow{2}{*}{Weighted average} \\ \cline{4-8}
\multicolumn{3}{l |}{} & '12 & '13 & '14 & '15 & '16 & & \\
\midrule
\multicolumn{10}{l}{Clustering} \\
\midrule
Algorithm & Source & $k$ &  \multicolumn{7}{c}{} \\ \cline{1-10}
\multirow{8}{=}{k-Means} & \multirow{4}{=}{English Vocabulary} & 100 & $44.12$ & $35.77$ & $42.55$ & $50.11$ & $53.24$ & $45.16\pm6.10$ & $44.39\pm4.90$ \\
& & 1000 & $49.02$ & $39.74$ & $48.05$ & $55.73$ & $57.66$ & $50.04\pm6.35$ & $49.40\pm5.21$ \\
& & 10000 & $52.62$ & $42.69$ & $52.76$ & $59.73$ & $61.48$ & $53.86\pm6.63$ & $53.27\pm5.46$ \\
& & 25000 & $53.36$ & $43.28$ & $53.80$ & $60.72$ & $62.35$ & $54.70\pm6.75$ & $54.12\pm5.60$ \\ \cline{2-10}
& \multirow{4}{=}{English words (top 100000)} & 100 & $49.84$ & $38.85$ & $48.81$ & $56.37$ & $55.76$ & $49.93\pm6.32$ & $49.71\pm5.37$ \\
& & 1000 & $53.85$ & $42.64$ & $53.31$ & $60.38$ & $60.34$ & $54.10\pm6.49$ & $53.85\pm5.50$ \\
& & 10000 & $56.19$ & $45.77$ & $57.33$ & $63.49$ & $63.96$ & $57.35\pm6.59$ & $56.96\pm5.52$ \\
& & 25000 & $56.31$ & $45.75$ & $57.65$ & $64.02$ & $64.47$ & $57.64\pm6.79$ & $57.22\pm5.70$ \\ \cline{1-10}
\multirow{4}{=}{Spherical k-Means} & \multirow{4}{=}{Frequent words (top 50000)} & 100 & $53.47$ & $42.82$ & $54.53$ & $61.66$ & $61.93$ & $54.88\pm6.97$ & $54.42\pm5.87$ \\
& & 1000 & $56.88$ & $47.37$ & $59.18$ & $66.52$ & $65.89$ & $59.17\pm6.98$ & $58.62\pm5.98$ \\
& & 10000 & $58.80$ & $49.74$ & $62.21$ & $69.55$ & $69.05$ & $61.87\pm7.31$ & $61.18\pm6.30$ \\
& & 25000 & $58.91$ & $49.33$ & $62.19$ & $69.89$ & $69.02$ & $61.87\pm7.50$ & $61.22\pm6.48$ \\
\midrule
\multicolumn{10}{l}{Identity matrix} \\
\midrule
\multicolumn{3}{c |}{} & $61.31$ & $51.21$ & $\mathbf{67.47}$ & $\mathbf{72.90}$ & $\mathbf{73.88}$ & $\mathbf{65.35}\pm8.37$ & $\mathbf{64.55}\pm7.20$ \\
\midrule
\multicolumn{10}{l}{Multivariate analysis} \\ 
\midrule
Algorithm& \multicolumn{2}{Z |}{Source} &  \multicolumn{7}{c}{} \\ \cline{1-10}
\multirow{2}{*}{PCA} &  \multicolumn{2}{Z |}{English vocabulary} & $\mathbf{61.42}$ & $51.36$ & $66.44$ & $72.74$ & $73.72$ & $65.14\pm8.21$ & $64.32\pm7.00$ \\
& \multicolumn{2}{Z |}{Frequent words (top 100000)} & $60.03$ & $\mathbf{51.96}$ & $66.36$ & $72.39$ & $73.25$ & $64.80\pm7.99$ & $63.81\pm6.93$ \\
\midrule
\multicolumn{10}{l}{} \\
\midrule
\multicolumn{3}{l |}{Average} & $54.99\pm4.88$ & $43.65\pm4.03$ & $54.36\pm5.63$ & $63.75\pm6.65$ & $64.40\pm6.22$ & & \\
\bottomrule
\end{tabularx}
\end{center}
\label{tab:ftsummary}
\end{sidewaystable}

\begin{sidewaystable}[htp]
\caption{Complete experimental results (measured as Spearman's correlation with human judgment) of the SFBoW model on the STS tasks using a Sent2Vec matrix as word embedding source. On the rightmost columns are reported an unweighted and a weighted per-model summary over the tasks, while on the bottom row is reported a per-task summary over the models, finally on the bottom-right corner are the unweighted and weighted per-word embedding matrix statistics. Bold values represent the highest score in the corresponding column.}
\begin{center}
\begin{tabularx}{\textwidth}{B | B | S | lllll | c | c}
\toprule
\multicolumn{3}{l |}{} & \multicolumn{7}{c}{Results (Spearman's $\rho$)} \\ \cline{4-10}
\multicolumn{3}{l |}{Universe matrix} &\multicolumn{5}{c |}{STS} & \multirow{2}{*}{Average} & \multirow{2}{*}{Weighted average} \\ \cline{4-8}
\multicolumn{3}{l |}{} & '12 & '13 & '14 & '15 & '16 & & \\
\midrule
\multicolumn{10}{l}{Clustering} \\
\midrule
Algorithm & Source & $k$ &  \multicolumn{7}{c}{} \\ \cline{1-10}
\multirow{8}{=}{k-Means} & \multirow{4}{=}{English vocabulary} & 100 & $51.14$ & $37.95$ & $54.58$ & $60.81$ & $58.30$ & $52.56\pm8.01$ & $52.48\pm6.97$ \\
& & 1000 & $51.94$ & $39.14$ & $56.24$ & $62.88$ & $61.09$ & $54.26\pm8.47$ & $53.95\pm7.35$ \\
& & 10000 & $53.26$ & $42.33$ & $59.00$ & $65.43$ & $62.86$ & $56.58\pm8.22$ & $56.16\pm7.27$ \\
& & 25000 & $54.76$ & $43.77$ & $61.39$ & $67.01$ & $64.93$ & $58.37\pm8.40$ & $57.93\pm7.44$ \\ \cline{2-10}
& \multirow{4}{=}{Frequent words (top 100000)} & 100 & $51.65$ & $37.00$ & $54.51$ & $59.98$ & $58.78$ & $52.38\pm8.25$ & $52.38\pm7.08$ \\
& & 1000 & $52.96$ & $40.84$ & $58.51$ & $64.20$ & $62.09$ & $55.72\pm8.36$ & $55.43\pm7.33$ \\
& & 10000 & $53.85$ & $41.98$ & $60.08$ & $66.01$ & $63.83$ & $57.15\pm8.63$ & $56.76\pm7.61$ \\
& & 25000 & $54.76$ & $43.77$ & $61.39$ & $67.01$ & $64.93$ & $58.37\pm8.40$ & $57.93\pm7.44$ \\ \cline{1-10}
\multirow{4}{=}{Spherical k-Means} & \multirow{4}{=}{Frequent words (top 50000)} & 100 & $50.83$ & $38.45$ & $54.94$ & $59.92$ & $59.25$ & $52.68\pm7.83$ & $52.44\pm6.74$\\
& & 1000 & $53.38$ & $41.06$ & $58.17$ & $63.46$ & $62.22$ & $55.66\pm8.11$ & $55.39\pm7.03$ \\
& & 10000 & $54.95$ & $43.71$ & $61.12$ & $66.72$ & $64.94$ & $58.29\pm8.33$ & $57.87\pm7.33$ \\
& & 25000 & $\mathbf{55.22}$ & $43.48$ & $61.53$ & $67.13$ & $\mathbf{65.22}$ & $\mathbf{58.52}\pm8.54$ & $\mathbf{58.12}\pm7.52$ \\
\midrule
\multicolumn{10}{l}{Identity matrix} \\
\midrule
\multicolumn{3}{c |}{} & $52.80$ & $\mathbf{44.05}$ & $\mathbf{62.23}$ & $\mathbf{68.63}$ & $63.96$ & $58.33\pm8.80$ & $57.75\pm8.16$\\
\midrule
\multicolumn{10}{l}{Multivariate analysis} \\ 
\midrule
Algorithm& \multicolumn{2}{Z |}{Source} &  \multicolumn{7}{c}{} \\ \cline{1-10}
\multirow{2}{*}{PCA} &  \multicolumn{2}{Z |}{English vocabulary} & $52.51$ & $43.09$ & $61.72$ & $67.84$ & $63.26$ & $57.68\pm8.83$ & $57.18\pm8.14$ \\
& \multicolumn{2}{Z |}{Frequent words (top 100000)} & $52.84$ & $43.50$ & $62.04$ & $67.88$ & $63.83$ & $58.02\pm8.77$ & $57.48\pm8.06$ \\
\midrule
\multicolumn{10}{l}{} \\
\midrule
\multicolumn{3}{c |}{Average} & $53.12\pm1.34$ & $41.58\pm2.32$ & $59.24\pm2.78$ & $64.99\pm2.86$ & $62.64\pm2.24$ & & \\
\bottomrule
\end{tabularx}
\end{center}
\label{tab:s2vsummary}
\end{sidewaystable}

\end{document}